\documentclass[acmsmall,screen]{acmart}
\usepackage[normalem]{ulem}

\AtBeginDocument{%
  }

\begin{document}

\title{Deep Learning for Sports Video Event Detection: Tasks, Datasets, Methods, and Challenges}

\author{Hao Xu}
\affiliation{%
  \institution{School of Information Technology, Deakin University}
  \city{Melbourne}
  \country{Australia}
}
\email{august.xu@research.deakin.edu.au}

\author{Arbind Agrahari Baniya}
\affiliation{%
  \institution{School of Information Technology, Deakin University}
  \city{Melbourne}
  \country{Australia}
}
\email{arbind.baniya@deakin.edu.au}

\author{Sam Wells}
\affiliation{%
  \institution{Paralympics Australia}
    \city{Melbourne}
  \country{Australia}
}
\email{sam.wells@paralympic.org.au}

\author{Mohamed Reda Bouadjenek}
\affiliation{%
  \institution{School of Information Technology, Deakin University}
    \city{Melbourne}
  \country{Australia}
}
\email{reda.bouadjenek@deakin.edu.au}

\author{Richard Dazeley}
\affiliation{%
  \institution{School of Information Technology, Deakin University}
    \city{Melbourne}
  \country{Australia}
}
\email{richard.dazeley@deakin.edu.au}

\author{Sunil Aryal}
\affiliation{%
  \institution{School of Information Technology, Deakin University}
  \city{Melbourne}
  \country{Australia}
}
\email{sunil.aryal@deakin.edu.au}

\renewcommand{\shortauthors}{Xu et al.}

\begin{abstract}
Video event detection has become a cornerstone of modern sports analytics, powering automated performance evaluation, content generation, and tactical decision-making. Recent advances in deep learning have driven progress in related tasks such as Temporal Action Localization (TAL), which detects extended action segments; Action Spotting (AS), which identifies a representative timestamp; and Precise Event Spotting (PES), which pinpoints the exact frame of an event. Although closely connected, their subtle differences often blur the boundaries between them, leading to confusion in both research and practical applications.
Furthermore, prior surveys either address generic video event detection or broader sports video tasks, but largely overlook the unique temporal granularity and domain-specific challenges of event spotting. In addition, most existing sports video surveys focus on elite-level competitions while neglecting the wider community of everyday practitioners.
This survey addresses these gaps by: (i) clearly delineating TAL, AS, and PES and their respective use cases; (ii) introducing a structured taxonomy of state-of-the-art approaches—including temporal modeling strategies, multimodal frameworks, and data-efficient pipelines tailored for AS and PES; and (iii) critically assessing benchmark datasets and evaluation protocols, highlighting limitations such as reliance on broadcast-quality footage and metrics that over-reward permissive multi-label predictions. By synthesizing current research and exposing open challenges, this work provides a comprehensive foundation for developing temporally precise, generalizable, and practically deployable sports event detection systems for both the research and industry communities.
\end{abstract}

\begin{CCSXML}
<ccs2012>
   <concept>
       <concept_id>10010147.10010178.10010224.10010225.10010228</concept_id>
       <concept_desc>Computing methodologies~Activity recognition and understanding</concept_desc>
       <concept_significance>500</concept_significance>
       </concept>
   <concept>
       <concept_id>10010147.10010178.10010224.10010245.10010248</concept_id>
       <concept_desc>Computing methodologies~Video segmentation</concept_desc>
       <concept_significance>300</concept_significance>
       </concept>
   <concept>
       <concept_id>10010147.10010178.10010224.10010225.10010230</concept_id>
       <concept_desc>Computing methodologies~Video summarization</concept_desc>
       <concept_significance>500</concept_significance>
       </concept>
 </ccs2012>
\end{CCSXML}

\ccsdesc[500]{Computing methodologies~Activity recognition and understanding}
\ccsdesc[300]{Computing methodologies~Video segmentation}
\ccsdesc[500]{Computing methodologies~Video summarization}
\keywords{Event Detection, Temporal Action Localization, Sports Videos}


\maketitle

\section{Introduction\label{sec:introduction}}
Sports represent one of the largest global markets, projected to reach 599.9 billion US dollars by 2025 and 826 billion US dollars by 2030, with a compound annual growth rate of 6.6\% \cite{kumar2021global}. Beyond its economic impact in industries such as media, marketing, and apparel, sports are fundamentally focused on athletic performance, where optimising player efficiency, refining game strategies, and enhancing fan engagement are critical. 
The rise of sports analytics, the systematic collection, processing, and analysis of performance data, has enabled data-driven decision-making, leading to fundamental changes in strategy. For instance, basketball has seen an increased reliance on three-point shooting, guided by predictive models that estimate expected points from various court locations \cite{morgulev2018sports}.

Within this context, video event detection has emerged as a fundamental yet challenging task in sports analytics. Accurate identification of key moments, such as corner kicks in soccer, rally conclusions in racket sports, or scoring events across disciplines, provides critical insight to coaches and athletes, supporting more effective performance analysis and tactical decision-making. Moreover, event detection benefits downstream applications by filtering non-playing segments, optimising computational resources for subsequent tasks such as object tracking, and enabling automated highlight generation for commercial broadcasting and fan engagement.

\begin{figure}[h]
    \centering
    \includegraphics[width=0.6\linewidth]{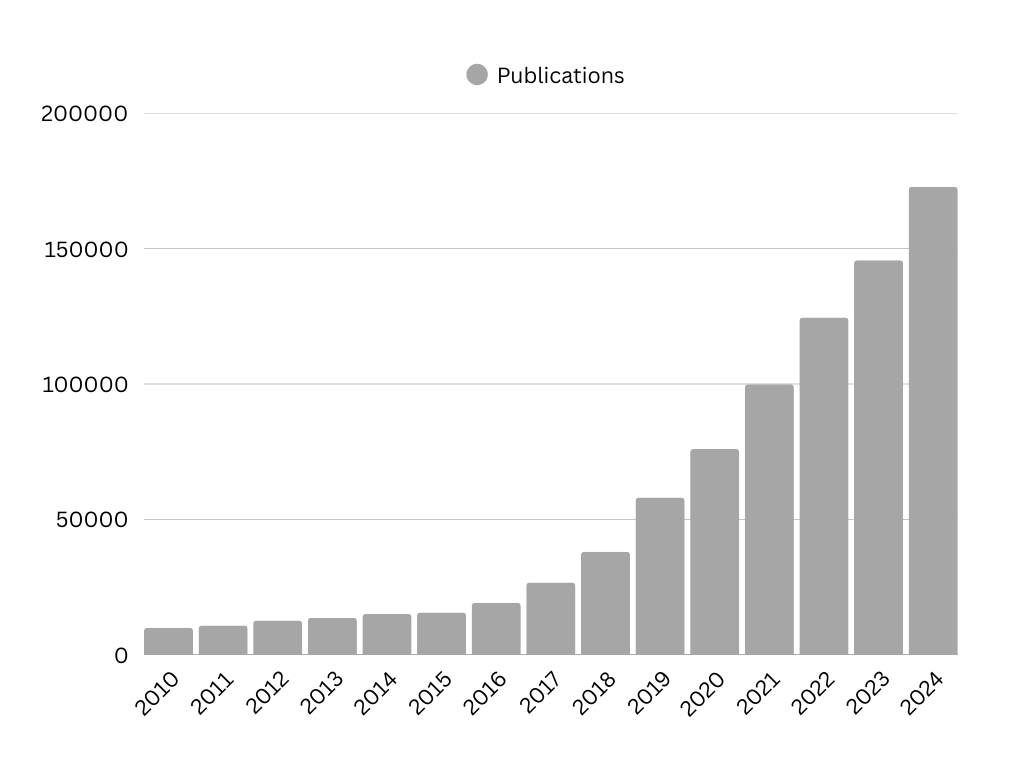}
    \caption{Annual publication count from 2010 to 2024 based on a Scopus keyword search for "sports" AND "deep learning". }
    \Description{A line chart showing the number of publications per year from 2010 to 2024 with the keywords "sports" and "deep learning" in Scopus. The trend shows very few publications before 2015, followed by steady growth, and a sharp increase after 2018, reaching the highest counts in recent years.}
    \label{fig:trend_chart}
\end{figure}

Computer Vision (CV) has played a pivotal role in advancing sports video analysis, enabling automated player tracking, action recognition, tactical analysis, injury prevention, and officiating through advanced visual data processing \cite{naik2022comprehensive, zhao2023survey, ghosh2023sports, wu2022survey, dindorf2022conceptual, host2022overview}. While traditional CV methods laid the foundation for many applications \cite{ghosh2023sports, thomas2017computer}, they heavily relied on handcrafted features and struggled with dynamic environments, occlusions, and real-time constraints. 
Recent breakthroughs in Deep Learning (DL), particularly through Convolutional Neural Networks (CNNs) and Transformer architectures, have significantly enhanced the accuracy, efficiency, and scalability of sports video analysis tasks \cite{ghosh2023sports}. These modern techniques now power real-time object detection, robust player tracking such as \cite{tarashima2023widely, huang2019tracknet, zhu2020deformable}, pose estimation \cite{etaat2025latte}, and fine-grained event recognition \cite{he2024vistecvideomodelingsports, lai2024facts}, fundamentally reshaping the landscape of sports analytics.

To further illustrate the growing interest in this domain, we compiled a publication trend analysis using Scopus, based on keyword searches related to "sports" and "deep learning." The resulting bar chart (Figure~\ref{fig:trend_chart}) shows a clear year-on-year increase in the number of relevant publications from 2010 to 2024, reinforcing the expanding research momentum in this field.

As video event detection in sports continues to advance with deep learning-based computer vision, the task has evolved into three closely related formulations. \textit{Temporal Action Localization (TAL)} detects extended temporal segments of an action (e.g., the duration of a soccer corner kick); \textit{Action Spotting (AS)} identifies a single representative keyframe of an event (e.g., the release of a basketball shot); and \textit{Precise Event Spotting (PES)} imposes stricter temporal accuracy requirements than AS by pinpointing the exact timestamp of an event with frame-level precision (e.g., the instant a table tennis ball bounces). Despite their similarities, these formulations often cause confusion among both academic and industry readers, as the subtle distinctions make it unclear which task is most appropriate for a given application.

Sports video event detection also introduces unique challenges that distinguish it from generic action understanding. Events are often brief and demand frame-level precision, in contrast to the longer temporal windows common in TAL benchmarks. Frequent occlusions from players or equipment, rapid object motion, and small target sizes further complicate detection \cite{xu2025totnet, huang2019tracknet}. In practical deployments, analysts typically work with monocular broadcast footage or resource-constrained capture environments, limiting the availability of multi-view or high-resolution data \cite{xu2025multi}. Moreover, evaluation protocols originally designed for broader action recognition frequently overlook the strict temporal accuracy required in sports, sometimes rewarding overly permissive multi-label predictions that provide little practical value. 

Previous studies, such as those by Ghosh et al. \cite{ghosh2023sports}, broadly survey AI applications in sports analytics but do not specifically address DL-based CV methods. Similarly, Thomas et al. \cite{thomas2017computer} focus primarily on traditional CV approaches designed for multi-camera systems. In contrast, our survey specifically targets deep learning-based approaches—including recent advances in\textbf{ CNNs and Vision Transformers}—for event detection tasks within \textbf{monocular video contexts}, enhancing its relevance for real-world deployment.
Other comprehensive surveys by Naik et al. \cite{naik2022comprehensive}, Karoline et al. \cite{seweryn2023survey}, Zhao et al. \cite{zhao2023survey}, Kamble et al. \cite{kamble2019ball}, Wu et al. \cite{wu2022survey}, and Yin et al. \cite{yin2024survey} extensively review CV methods across various sports types and analytics tasks, including object tracking and action recognition. However, they do not specifically address video event detection across sports disciplines. The most closely related work by Hu et al. \cite{hu2024overview} provides an extensive overview of TAL methods, but it does not cover the increasingly critical tasks of AS and PES, nor is it specifically focused on the sports domain.

Motivated by these gaps and challenges, our objective is to consolidate recent progress in sports video event detection with a particular focus on \textbf{DL approaches}, establish clear task definitions, and critically examine methods, datasets, and evaluation practices in the context of real-world sports analytics, while also providing insights through in-depth discussions of open challenges and future directions.

To this end, our survey makes the following key contributions:

\begin{itemize}
    \item \textbf{Task Definitions and Distinctions:} We formally define and differentiate the three central event detection tasks in sports videos—TAL, AS, and PES—highlighting their objectives, annotation schemes, and relevance to sports scenarios requiring different levels of temporal precision.
    
    \item \textbf{Methodological Taxonomy:} We propose a taxonomy of deep learning approaches for AS and PES, reviewing temporal modeling methods, multi-model based methods, and data-efficient frameworks built on convolutional, recurrent, and transformer models.
    
    \item \textbf{Datasets and Evaluation Protocols:} We summarize the benchmark datasets and evaluation metrics used across TAL, AS, and PES, and critically assess their suitability for real-world deployment. In particular, we highlight limitations related to confidence thresholding and multi-label predictions, while also discussing potential solutions.  

    \item \textbf{Insights and Future Directions:} We discuss open challenges such as poor generalization across sports, limited supervision strategies, and unrealistic evaluation schemes, and propose research directions toward more robust and deployable spotting models in practical application settings.
\end{itemize}

By aligning these contributions with the specific needs of sports event detection, our survey provides a focused and timely perspective that complements existing surveys in sports analytics and video understanding.

The remainder of the paper is organized as follows. Section~\ref{sec:Definition} provides clear definitions of the tasks involved in sports video event detection. Section~\ref{sec:VED} reviews existing methodologies in this domain. Sections~\ref{sec:datasets} and~\ref{sec:evaluation_metrics} introduce benchmark datasets and evaluation metrics commonly used for sports event detection. Section~\ref{sec:app} presents practical applications enabled by sports video event detection. Section~\ref{sec:challenges_directions} discusses key challenges and future research directions. Finally, Section~\ref{sec:conclusion} concludes the paper with a summary of insights and findings.

\section{Sports Event Detection}\label{sec:Definition}

In sports video analysis, three core tasks have emerged for temporal event detection: TAL, AS, and PES. Although related, they differ in output format, annotation granularity, and application focus (see Table~\ref{tab:task_comparison}). In this section, we clearly define each task and discuss their suitability for sports video event detection.

\begin{table}[h]
\centering
\renewcommand{\arraystretch}{1.4}
\caption{Comparison of Temporal Action Localisation (TAL), Action Spotting (AS), and Precise Event Spotting (PES).}
\resizebox{\linewidth}{!}{%
\begin{tabular}{lccc}
\toprule
\textbf{Aspect} & \textbf{TAL} & \textbf{AS} & \textbf{PES} \\
\midrule
\textbf{Output Type} & Temporal interval & Single key frame & Single key frame \\
\textbf{Annotation Format} & Start and end times & Single timestamp & Single timestamp \\
\textbf{Tolerance Window} & $\sim$1--5 seconds & 5--60 frames & 0--2 frames \\
\textbf{Best Suited For} & Long-duration actions & Ambiguous, fast-paced actions & Frame-accurate event detection \\
\textbf{Annotation Cost} & High & Medium & Medium \\
\textbf{Use Cases} & Long \& Continuous events & Sports highlight detection & Fine-grained critical events \\
\bottomrule
\end{tabular}}
\label{tab:task_comparison}
\end{table}

\subsection{Temporal Action Localization}
TAL—also referred to as Temporal Action Detection (TAD)—aims to detect and classify action segments within untrimmed videos. A common formulation builds on Temporal Action Proposal Generation (TAPG), which first identifies candidate temporal regions likely to contain actions and then assigns class labels \cite{lin2018bsn, lin2019bmn, hu2024overview}. This is typically achieved in two stages: (i) proposal generation, where potential action boundaries are suggested, and (ii) classification and refinement, where those proposals are labeled and their temporal boundaries adjusted.  

Originally developed for generic activity understanding on datasets such as ActivityNet \cite{caba2015activitynet} and THUMOS \cite{Idrees_2017}, TAL methods have since been adapted to sports because of their ability to model temporal structure over extended sequences. The main challenge lies in accurately predicting start and end boundaries, particularly for fine-grained or short-duration actions, whereas classification within well-defined proposals is comparatively more reliable \cite{hu2024overview}.  

In the context of sports, TAL is effective for analyzing long or continuous actions such as rallies in racket sports or set plays in team games. However, it is far less suitable for overlapping or instantaneous events—such as ball bounces or racket–ball contacts—that require frame-level precision. Consequently, TAL is best applied to coarse temporal segmentation tasks, including highlight generation and the detection of extended play phases.

\begin{figure}[htbp]
    \centering
    \includegraphics[width=1\linewidth]{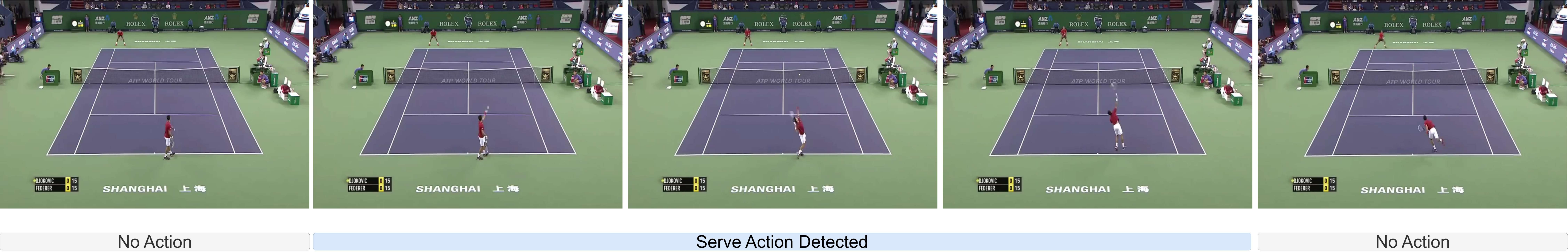}
    \caption{Example of Temporal Action Localization: in tennis, the full serve motion is annotated as a time interval (blue bar).}
    \Description{A timeline visualization of a tennis video showing how the full serve motion is labeled as a continuous time interval. The annotated segment is highlighted with a blue bar aligned to the serve action.}
    \label{fig:TAL_Example}
\end{figure}

\subsection{Action Spotting}
AS, introduced by Giancola et al. \cite{giancola2018soccernet}, was proposed to overcome the difficulty of annotating precise action boundaries in sports videos, where events are often rapid, overlapping, or continuous. Instead of labeling start and end times—which can be subjective and inconsistent—AS represents each event with a single timestamp, referred to as the “spotting point.” The goal of AS is therefore to predict the coarse frame in which an event occurs, rather than its full temporal extent.  

This task was first introduced on the SoccerNet dataset \cite{giancola2018soccernet}, where events such as goals, substitutions, and cards are inherently ambiguous in terms of their exact boundaries. To account for this ambiguity, predictions are evaluated within a relatively wide tolerance window (typically $\pm$50 frames). This formulation greatly simplifies annotation, reduces subjectivity, and enables the efficient construction of large-scale benchmarks such as SoccerNet and its extensions \cite{deliege2021soccernet}.  

Both TAL and AS aim to localize events temporally, but their priorities differ. TAL captures interval-level segments, making it effective for long, structured actions. AS, by contrast, trades interval precision for a more scalable and annotation-friendly framework, making it particularly well suited for rapid or ambiguous sports actions such as passes, shots, or fouls in soccer. The AS formulation also aligns closely with downstream applications such as match summarization and key-stat reporting, where identifying the representative moment of an action is often more useful than modeling its full duration.

\subsection{Precise Event Spotting}
PES, first proposed by Hong et al. \cite{hong2021video}, extends the AS formulation by enforcing strict frame-level precision. While AS evaluates predictions within a wide tolerance window (e.g., $\pm$50 frames), many sports events require far greater accuracy. For instance, tennis ball bounces or figure skating landings occur within only a few frames and must be identified precisely to provide valuable information for analysts and coaches. By tightening the temporal tolerance of AS, PES was introduced as a more suitable task for fine-grained sports scenarios.  

\begin{figure}[htbp]
    \centering
    \includegraphics[width=1\linewidth]{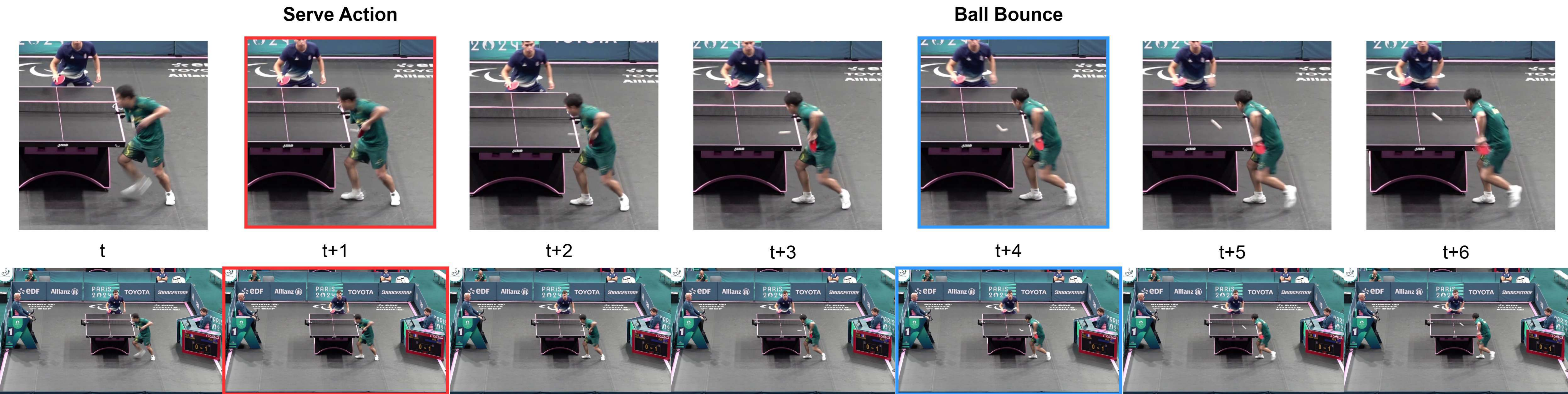}
    \caption{Example of Precise Event Spotting: in table tennis, the moment a player contacts the ball during a serve (red) or when a ball bounces on the table (blue).}
    \Description{A timeline visualization of a table tennis rally showing precise event spotting. Individual moments are marked: red dots indicate the exact frame when the player contacts the ball during a serve, while blue dots indicate the exact frame when the ball bounces on the table.}
    \label{fig:PES_Example}
\end{figure}

The need for such precision is further supported by feedback from Table Tennis Australia, where detecting ball contacts or bounce points often requires localization within under 100 milliseconds. Errors of even 1–2 frames can result in missing decisive events (see Figure~\ref{fig:PES_Example}).  

Because of these requirements, PES is increasingly being adopted by the community as the latest evolution of event detection tasks. Its high temporal fidelity makes it essential for applications such as biomechanics, officiating support, key-stat summarization, and highlight generation with frame-level accuracy. Recent datasets have already embraced this stricter formulation, including SoccerNet-v2 \cite{deliege2021soccernet}, tennis \cite{hong2021video}, and table tennis \cite{xu2025multi}.

\section{Video Event Detection}\label{sec:VED}
In this section, we provide an overview of video event detection methods in sports. We begin by describing the foundational general-purpose approaches that shaped many of the earliest sports video event detection pipelines. Next, we present a comprehensive review of methods specifically developed for sports, with a focus on the recent emergence of AS and PES. These tasks have been introduced to address the unique challenges of detecting fine-grained events in fast-paced and often ambiguous sports scenarios.

\subsection{Foundations of Temporal Action Localization}
\begin{figure*}[htbp]
    \centering
    \includegraphics[width=1.0\linewidth]{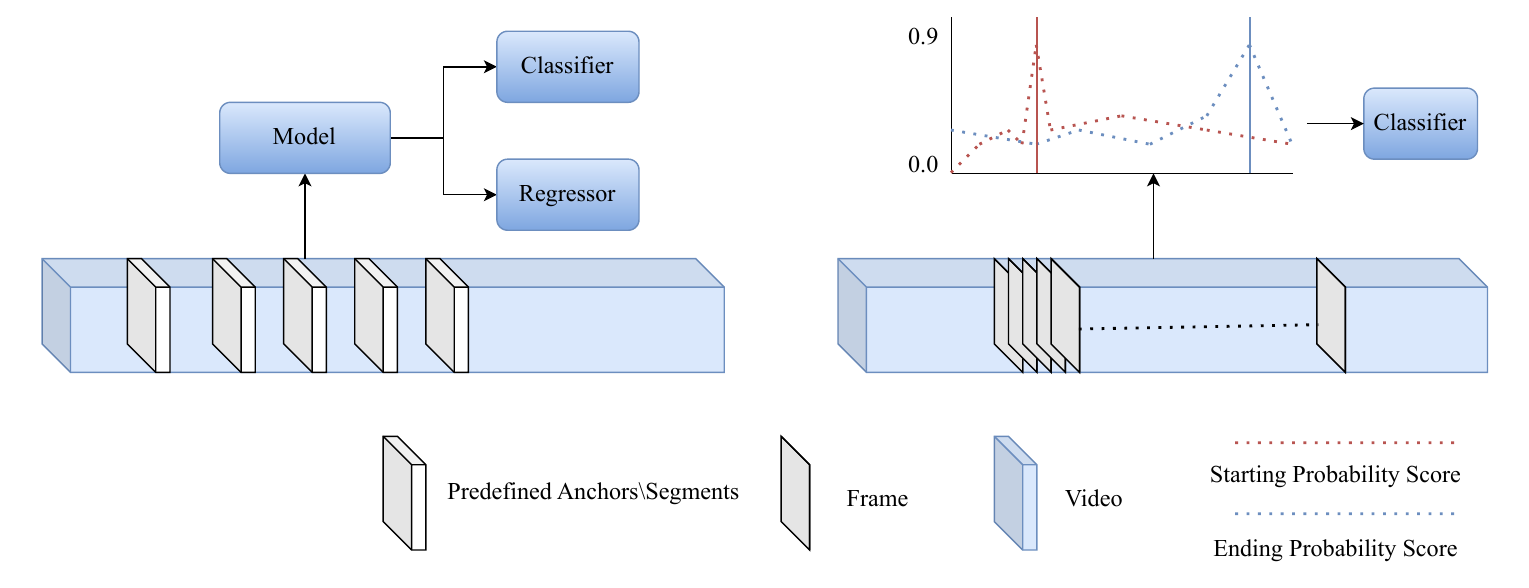}
    \caption{Comparison of Global-to-Local (GTL, left) and Local-to-Global (LTG, right) approaches. GTL classifies predefined temporal anchors as action or background, followed by regression to refine time intervals. LTG predicts per-frame start and end probabilities to define action boundaries, which are then classified.}
    \Description{A side-by-side schematic comparison of two approaches for temporal action localization. The Global-to-Local approach shows temporal anchors classified as action or background, followed by regression to refine interval boundaries. The Local-to-Global approach shows frame-wise start and end probabilities that define action intervals, which are then assigned to classes.}
    \label{fig:GTL_LTG_Comparsion}
\end{figure*}

Although our focus is on sports event detection, many current methods are rooted in advances from generic TAL. TAL methods generally fall into two paradigms: Global-to-Local (GTL), which generates proposals from predefined anchors or sliding windows, and Local-to-Global (LTG), which predicts per-frame start, end, and actionness probabilities before combining them into segments (Figure~\ref{fig:GTL_LTG_Comparsion}).

Early GTL approaches such as TURN \cite{gao2017turn} and CTAP \cite{gao2018ctap} relied on sliding-window proposals, offering strong recall but limited boundary precision. In contrast, LTG methods shifted the field toward boundary-sensitive modeling. The Boundary-Sensitive Network (BSN) \cite{lin2018bsn} was the first to predict frame-level start, end, and actionness scores, generating high-quality proposals with fewer candidates. Building on this idea, BMN \cite{lin2019bmn} introduced a boundary-matching mechanism to efficiently evaluate densely distributed proposals via a two-dimensional confidence map. Later refinements, such as BSN++ \cite{su2021bsn++}, incorporated completeness modeling and global-local fusion to improve proposal quality, while TCANet \cite{qing2021temporal} and BCNet \cite{yang2022temporal} leveraged context aggregation and attention mechanisms to further enhance boundary accuracy.

These advances established the technical foundation for event detection in sports. However, their reliance on extended temporal intervals limits applicability to fast, fine-grained events, as seen in TAL. For instance, while proposal-based methods can capture rallies in racket sports, they often fail to localize instantaneous actions such as ball bounces or racket–ball contacts. This limitation motivated the development of AS, which simplifies annotations to single timestamps, and PES, which enforces frame-level accuracy. In the following sections, we review methods explicitly designed for these tasks in sports video event detection.

\subsection{Sports Video Event Detection}

While general TAL-based methods laid the foundation for video event detection, their reliance on coarse temporal intervals limits applicability in sports. Consequently, research has increasingly shifted toward AS and PES, which emphasize frame-level precision. In this section, we review methodologies explicitly developed for these two tasks in sports video event detection.

\begin{figure}[htbp]
    \centering
    \includegraphics[width=1\linewidth]{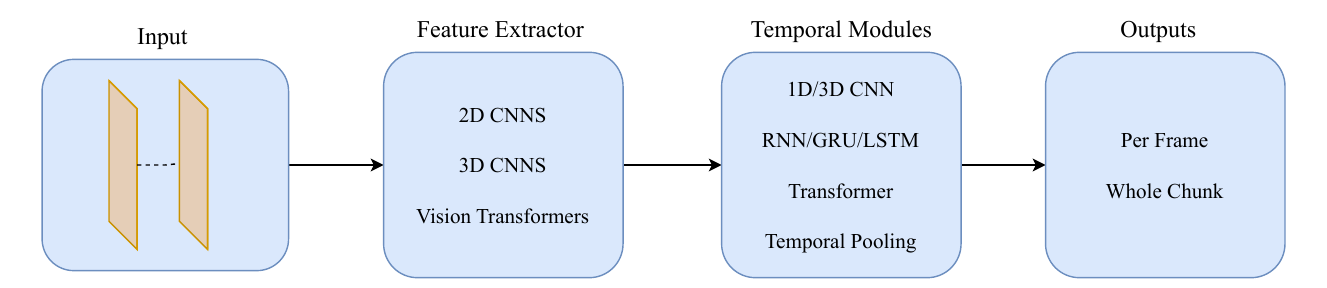}
    \caption{Typical workflow of AS/PES models: an input video clip composed of multiple frames is first processed by a feature extractor (2D or 3D), followed by temporal modules to capture temporal dependencies. The final output can be frame-level predictions or clip-level classifications, depending on the task requirements.}
    \Description{A diagram illustrating the standard pipeline for action spotting and precise event spotting models. A video clip of multiple frames is passed through a feature extractor, then temporal modeling modules, and finally produces either per-frame predictions or clip-level classifications depending on the task.}
    \label{fig:AS_PES_Pipeline}
\end{figure}

Although AS and PES differ in evaluation metrics and frame-level precision requirements, many detection models are applicable to both; we show a typical architecture workflow in Figure
\ref{fig:AS_PES_Pipeline}. 

We also categorize existing approaches according to their underlying architectural strategies:

\begin{itemize}
    \item \textbf{Temporal Modeling Methods} operate on frame-level or chunk-level features extracted from pretrained visual models to capture temporal structure within a window. These approaches range from (i) simple pooling-based aggregation (e.g., mean, max pooling, or NetVLAD++) that condenses temporal information for classification, to (ii) learnable sequence encoders (e.g., 1D/3D CNNs, RNNs, Transformers) that explicitly model dependencies across frames, and (iii) frame-aware architectures designed to capture subtle differences at the frame level for precise event localization in spotting tasks, all of them following by a classifier to classify the event either in frame level or chunk level.
    \item \textbf{Multi-Modal Fusion Methods} integrate additional modalities beyond visual signals—such as audio cues (e.g., game sounds, whistles, crowd reactions) or textual data (e.g., commentary transcripts)—to provide complementary context and improve event detection performance.  
    
    \item \textbf{Data-Efficient Learning Approaches} aim to reduce reliance on large-scale manual annotations by leveraging strategies such as semi-supervised learning, self-supervised pretraining, active learning, or knowledge distillation.

\end{itemize}

Table \ref{tab:as_pes_summary} provides an overview of the methods discussed in this section, along with their evaluation performance on the SoccerNet benchmark where available. The table also indicates the taxonomy category of each method, offering a concise comparison across different approaches.

\begin{table*}[t]
\centering
\renewcommand{\arraystretch}{1.1}
\caption{Summary of methods for AS and PES. Performance is grouped under the Test and Challenge sets. All results are reported on the SoccerNet Action Spotting dataset, where \textit{\uline{italicized entries}} indicate results from SoccerNet-v1 \cite{giancola2018soccernet}, and plain text entries are from SoccerNet-v2 \cite{deliege2021soccernet}. Bold numbers indicate the highest scores in each column. A \checkmark in the last column means the method was evaluated on datasets beyond SoccerNet. C.D. Eval means cross-dataset evaluation.}
\resizebox{\textwidth}{!}{%
\begin{tabular}{l|c|l|c|cc|cc|c}
\toprule
\textbf{Method} & \textbf{Year} & \textbf{Category} & \textbf{Parameter Size} &
\multicolumn{2}{c|}{\textbf{Test Set}} &
\multicolumn{2}{c|}{\textbf{Challenge Set}} &
\textbf{C.D Eval.} \\
\cmidrule(lr){5-6} \cmidrule(lr){7-8}
& & & & \textbf{Tight} & \textbf{Loose} & \textbf{Tight} & \textbf{Loose} & \\
\midrule
Giancola et al. \cite{giancola2018soccernet} & 2018 & Pooling-Based & -- & -- & 31.37 & -- & 30.74 & -- \\
Rongved et al. \cite{rongved2020real} & 2020 & Frame-Aware & -- & -- & \textit{\uline{56.3}} & -- & -- & -- \\
Vats et al. \cite{vats2020event} & 2020 & Temporal Encoder-Based & -- & -- & \textit{\uline{60.1}} & -- & -- & -- \\
CALF \cite{cioppa2020context} & 2020 & Pooling-Based & -- & -- & 41.61 & -- & 42.22 & -- \\
Vanderplaetsen et al. \cite{vanderplaetse2020improved} & 2020 & Multi-Modal Fusion & -- & -- & 39.90 & -- & -- & -- \\
NetVLAD++ \cite{giancola2021temporally} & 2021 & Pooling-Based & -- & -- & 53.40 & -- & 52.54 & -- \\
RMS-Net \cite{tomei2021rms} & 2021 & Frame-Aware & -- & -- & 63.49 & -- & 60.92 & -- \\
Zhou \cite{zhou2021feature} & 2021 & Pooling-Based & -- & 47.05 & 74.77 & 49.56 & 74.84 & -- \\
E2E-Spot \cite{hong2022spotting} & 2022 & Frame-Aware & 4.5M & -- & -- & 66.73 & 73.62 & \checkmark \\
Shi et al. \cite{shi2022action} & 2022 & Temporal Encoder-Based & -- & -- & 55.20 & -- & -- & -- \\
STE \cite{darwish2022ste} & 2022 & Temporal Encoder-Based & 2.3M & 58.29 & 71.58 & 58.71 & 70.49 & -- \\
SpotFormer \cite{cao2022spotformer} & 2022 & Temporal Encoder-Based & -- & 60.90 & 81.50 & -- & -- & -- \\
Soares et al. \cite{soares2022temporally} & 2022 & Temporal Encoder-Based & 8.9M & 65.07 & 78.59 & 68.33 & 78.06 & -- \\
Zhu et al. \cite{zhu2022transformer} & 2022 & Pooling-Based & -- & -- & -- & 52.04 & 60.86 & -- \\
ASTRA \cite{xarles2023astra} & 2023 & Multi-Modal Fusion & -- & -- & -- & 70.10 & 79.21 & -- \\
T-DEED \cite{xarles2024t} & 2024 & Frame-Aware & 16.4M & -- & -- & -- & -- & \checkmark \\
COMEDIAN \cite{denize2024comedian} & 2024 & Data-Efficient Learning & 29.1M & 73.10 & -- & 68.38 & 73.98 & -- \\
Tran et al. \cite{tran2024unifying} & 2024 & Frame-Aware & -- & 62.49 & 73.98 & \textbf{69.38} & \textbf{76.15} & \checkmark \\
Santra et al. \cite{santra2025precise}& 2025 & Frame-Aware& 6.46M & \textbf{73.74} & \textbf{79.11}& - & - & \checkmark\\
\bottomrule
\end{tabular}}
\label{tab:as_pes_summary}
\end{table*}

\subsubsection{Temporal Modeling Methods}
\begin{figure}[htbp]
    \centering
    \includegraphics[width=1\linewidth]{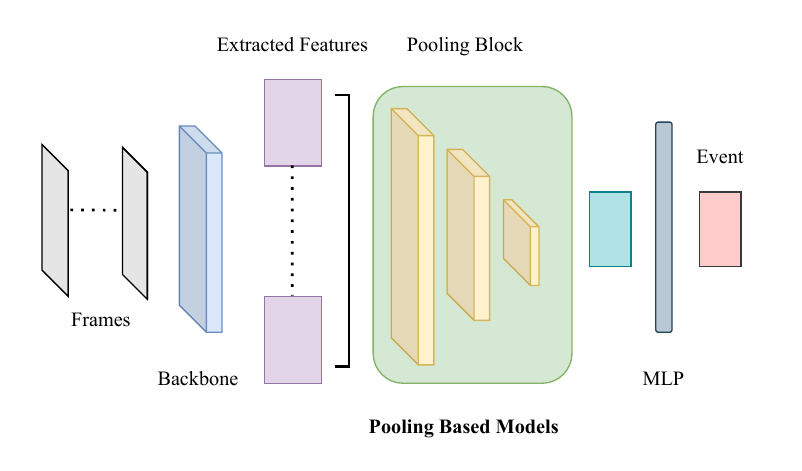}
    \caption{Pooling-based models for sports video event detection. Video frames are processed by a CNN backbone to extract features, which are then aggregated within a temporal window using pooling methods (e.g., mean, max, NetVLAD). The pooled representation is classified into event probabilities through the MLP.}
    \Description{A schematic diagram of a pooling-based sports video event detection model. Individual video frames are passed through a CNN backbone to obtain features, which are aggregated over a temporal window using pooling operations such as mean, max, or NetVLAD. The pooled feature vector is then fed into a multilayer perceptron that outputs event probabilities.}
    \label{fig:pooling_model}
\end{figure}

\paragraph{Pooling-Based} approaches typically adopt a sliding-window strategy, where videos are divided into fixed-length temporal segments containing a set of frames. Frame or chunk-level features are first extracted using backbone models such as 2D or 3D CNNs, and then aggregated over the temporal window using techniques such as average pooling, NetVLAD \cite{arandjelovic2016netvlad}, or the temporally-aware variant NetVLAD++ \cite{giancola2021temporally}. The aggregated representation is subsequently passed to a classifier to predict event labels. The overview of the pooling based models is shown in Figure \ref{fig:pooling_model}

One of the earliest works in this category is the SoccerNet baseline \cite{giancola2018soccernet}, which compared a range of pooling techniques—including mean, max, NetVLAD \cite{arandjelovic2016netvlad}, NetRVLAD \cite{arandjelovic2016netvlad}, NetFV, and SoftDBOW \cite{miech2017learnable} —on pre-extracted window features from C3D \cite{tran2015learning}, I3D \cite{carreira2017quo}, and ResNet \cite{he2016deep} to classify segments of soccer matches. The best performance was obtained by combining ResNet features with NetRVLAD pooling. Interestingly, 2D CNNs were found to outperform 3D CNNs in this setting. A possible explanation is that 3D CNNs already encode temporal dynamics during feature extraction, and further temporal aggregation through pooling may introduce redundancy or noise. In contrast, 2D CNNs primarily capture spatial information, allowing pooling strategies to more effectively extract complementary temporal cues.

Building on this direction, Rongved et al. \cite{rongved2020real} investigated the use of 3D ResNet \cite{hara2017learning} models pretrained on Kinetics-400 \cite{kay2017kinetics}, adapted from \cite{tran2018closer}. Their approach enhanced the ability to capture temporal dynamics, demonstrating that models pretrained on large-scale video action recognition datasets can be effectively transferred to event detection in sports. Specifically, the model was fed with input segments of 128 frames, and post-processing was applied using a moving average filter and non-maximum suppression (NMS) to reduce noise and prevent overlapping predictions of the same class.

Zhou et al. \cite{zhou2021feature} advanced this line of work by fine-tuning multiple action recognition backbones—including TPN \cite{yang2020temporal}, GTA \cite{he2020gta}, VTN \cite{neimark2021video}, irCSN \cite{tran2019video}, and I3D-Slow \cite{feichtenhofer2019slowfast}—on soccer video snippets. The combined features, when processed by NetVLAD++ \cite{giancola2021temporally}, achieved state-of-the-art performance on the SoccerNet benchmark. These results highlighted the effectiveness of ensemble learning in sports video event detection; however, the approach also raised efficiency concerns, particularly for real-time applications.

Zhu et al. \cite{zhu2022transformer} proposed a more efficient approach by employing a single multi-scale Vision Transformer (MViT) \cite{li2022mvitv2} for feature extraction on each proposal consisting of 16 frames. These frames were sampled at a stride of four from the original video, meaning that each proposal effectively covered 64 consecutive frames. The extracted features were then aggregated using NetVLAD++ pooling following by a fully connected layer to classify labels. This design achieves a balance between temporal modeling capacity and computational efficiency, making it well suited for deployment in resource-constrained settings.

One of the major challenges in sports video event detection is severe class imbalance: background (non-event) segments dominate most of the video, while meaningful events occupy only a small fraction. Standard loss functions such as cross-entropy typically neglect the contribution of frames surrounding an event, treating them as background. To address this, Cioppa et al. \cite{cioppa2020context} proposed a context-aware loss function that leverages temporal structure by dynamically weighting frames based on their proximity to annotated events. By adopting the smooth temporal weighting, it improved the baseline’s ability to focus on relevant cues and yielded a 12.8\% performance gain on SoccerNet-v1 \cite{giancola2018soccernet}. However, its effectiveness diminishes on denser datasets such as PES \cite{hong2022spotting}, where frame-level precision is critical.

Pooling-based methods have their merits: they are easy to implement, typically consisting of a CNN feature extractor combined with a temporal pooling mechanism. They are also computationally efficient, especially when compared to more complex temporal modeling approaches such as RNNs or Transformers. However, these advantages come with critical limitations.  

Most pooling-based approaches rely on generic CNN feature extractors that were not specifically designed for sports videos. This presents several challenges: the high frame rates of sports footage often result in adjacent frames that look very similar, while key details of interest (e.g., a tennis ball) are extremely small relative to the entire frame. Such conditions make it especially difficult for generic backbones to capture the fine-grained features needed for accurate event detection.  

Furthermore, temporal pooling itself discards sequential information. Even advanced pooling methods such as NetVLAD++ inevitably compress temporal dynamics into a fixed representation, which limits frame-level precision. This is particularly problematic in sports video detection, where accurate localization at the frame level is crucial. These limitations are also reflected in the AS task, helping to explain why recent research has shifted toward methods that emphasize fine-grained, frame-level precision, as exemplified by PES.

\begin{figure}[htbp]
    \centering
    \includegraphics[width=1\linewidth]{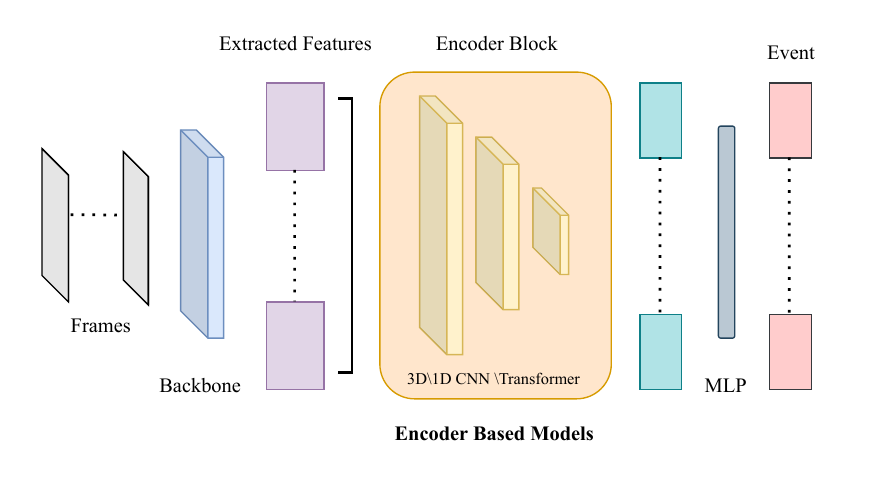}
    \caption{Illustration of encoder-based methods for AS and PES. Frame-level features are first extracted by a backbone and then processed by temporal encoders (e.g., 1D/3D CNNs, RNNs, Transformers) that preserve the sequence length and model dependencies across time. Predictions can then be made either at the segment level (after temporal pooling) or at the frame level for fine-grained event spotting.}
    \Description{A schematic diagram of encoder-based approaches for action spotting and precise event spotting. Frame-level features are extracted with a backbone and passed into temporal encoders such as 1D or 3D CNNs, RNNs, or Transformers, which preserve sequence length and capture temporal dependencies. The outputs can be aggregated for segment-level classification or retained per frame for fine-grained spotting.}
    \label{fig:encoder}
\end{figure}

\paragraph{Encoder Methods} enhance feature exploitation pipelines by replacing pooling operations with sequence models that explicitly capture temporal dependencies across frames. Examples include 1D CNNs, 3D CNNs \cite{tran2015learningspatiotemporalfeatures3d}, RNNs, and Transformers \cite{vaswani2017attention}, which enable richer contextual modeling. Unlike pooling-based approaches, these methods preserve the temporal dimension, allowing predictions to be made at the frame level and providing greater flexibility for fine-grained event localization. The overview of the encoder methods is shown in Figure \ref{fig:encoder}.

To address the variability in action duration, Vats et al. \cite{vats2020event} proposed a multi-tower 1D CNN architecture that processes input features at multiple temporal resolutions in parallel. Each tower uses different kernel sizes to capture short-term dynamics and longer-term dependencies, and their outputs are concatenated before final classification. Although the temporal dimension is ultimately collapsed—reflecting the coarse one-minute annotations of SoccerNet-v1 and NHL dataset \cite{vats2020event}, the multi-scale encoding provides richer intermediate representations, improving robustness across diverse sporting actions compared to single-resolution baselines.

Tomei et al. \cite{tomei2021rms} introduced RMS-Net, which formulates sports event detection in a manner similar to TAL. First, frame-level features are extracted from the entire video chunk. These features are then fused along the temporal dimension using 1D convolutions, followed by a max operation to remove the time axis. The resulting representation is processed by two output heads: a regression head that predicts temporal offsets and a classification head that assigns action labels.  
In addition, they proposed a novel data augmentation strategy motivated by the observation that the most informative visual cues often occur in frames immediately preceding or following an event. By randomly masking a portion of these frames, the model is forced to rely on either pre-event or post-event information, thereby improving robustness. This strategy yielded a 2.5 mAP improvement in evaluation, demonstrating the effectiveness of targeted temporal masking.

SpotFormer \cite{cao2022spotformer} further demonstrates the strength of sequence modeling approaches by fusing features from multiple pretrained backbones (e.g., VideoMAE \cite{tong2022videomae}, Swin Transformer \cite{liu2021swin}). The authors argue that different backbones capture complementary high-level spatiotemporal information, which benefits the temporal encoder. To combine these representations, features are first processed through isolated multilayer perceptrons (MLPs) and then concatenated along the channel dimension. The fused features are subsequently passed to a Transformer-based spotting head composed of several encoder blocks that model frame-wise interactions, followed by fully connected layers that predict per-frame action probabilities. This design enables frame-level predictions with high accuracy, but the model’s complexity makes it less practical for real-time deployment.

To address computational constraints, Darwish et al. \cite{darwish2022ste} introduced the Spatio-Temporal Encoder (STE), a lightweight architecture based on 1D convolutions and MLPs. Although the design is relatively simple, the model emphasizes efficiency, achieving competitive accuracy with substantially lower computational cost. Notably, STE can be trained entirely on CPUs, in contrast to most other methods that require GPUs, highlighting its practicality for deployment in resource-constrained or real-time sports analytics settings.

Soares et al. \cite{soares2022temporally} adapted an anchor-based detection framework—originally developed for TAL—to the AS domain. In their design, a video chunk is first passed through a feature extractor to obtain frame-level features, which are then reduced in dimensionality using a two-layer MLP. These features are processed by a trunk module that follows an encoder–decoder structure, where temporal information is progressively compressed and then restored. The trunk can be instantiated either as a 1D U-Net \cite{ronneberger2015u} or as a Transformer, enabling a trade-off between local boundary sensitivity and long-range context modeling. The processed features are finally passed through convolutional layers and two output heads: one for temporal offset regression and another for action classification. This design achieved strong results on the SoccerNet Challenge benchmark \cite{cioppa2024soccernet}. However, as with other anchor-based approaches, the reliance on pre-defined temporal scales remains a limitation, reducing adaptability to instantaneous events such as ball bounces in PES.

Shi et al. \cite{shi2022action} addressed the challenge of variable event durations by introducing a multi-scene encoding strategy. Instead of processing a fixed-length input, video chunks are segmented into similar-duration subsets, each handled by a dedicated Transformer encoder. This design enables the network to adapt its receptive field to both short-lived actions (e.g., passes, shots) and longer phases of play (e.g., build-up sequences), improving robustness across timescales. While effective, this approach comes at the cost of increased computational complexity due to maintaining multiple Transformer branches, making real-time deployment more challenging. Nevertheless, it highlights an important direction for spotting models—explicitly accounting for the highly diverse temporal granularity of sports events.

Taken together, these approaches illustrate the progression of sequence modeling in sports event detection—from early multi-scale convolutional encoders designed for coarse annotations, through hybrid encoder–decoder architectures with offset regression, to more recent Transformer-based spotting models that emphasize frame-level precision. While accuracy has steadily improved, trade-offs remain between temporal granularity, computational cost, and real-time applicability.

\begin{figure}[htbp]
    \centering
    \includegraphics[width=1\linewidth]{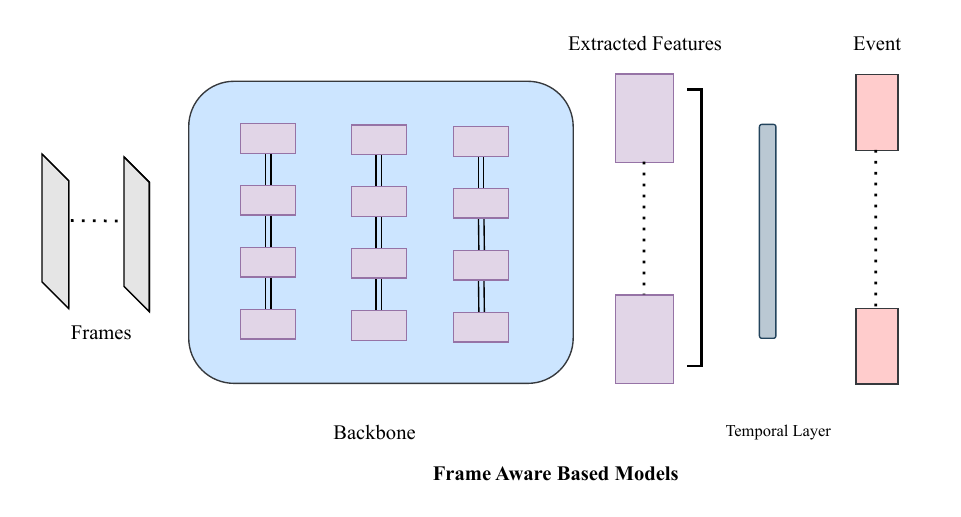}
    \caption{Overview of frame-aware models. Input frames are processed by a backbone that is modified by temporal modules (e.g., GSM, GRU, or Transformers) designed to enhance frame-level discriminability. The model outputs event probabilities for each frame, providing the strict temporal precision required for PES.}
    \Description{A schematic diagram of frame-aware models for precise event spotting. Video frames are passed through a backbone network combined with temporal modules such as gated shift modules, recurrent units, or transformers. The system outputs per-frame event probabilities, enabling fine-grained temporal precision in spotting tasks.}
    \label{fig:frame_aware}
\end{figure}

\paragraph{Frame-Aware} Models represent the most recent research direction, aiming to enhance spatiotemporal representation by directly modifying backbone architectures and temporal modeling to address the specific demands of sports video analysis. These approaches introduce frame-specific mechanisms that preserve the full temporal dimension, enabling true frame-level predictions and improving temporal discriminability for PES. Unlike pooling- and encoder-based methods, which primarily adapt architectures developed for generic video tasks, frame-aware models jointly learn low-level visual cues and high-level temporal semantics tailored to spotting, resulting in more accurate and temporally precise outcomes. The overview of the frame-aware models is shown in Figure \ref{fig:frame_aware}.

Hong et al. \cite{hong2022spotting} introduced E2E-Spot, the first frame-aware model, while also formally proposing the task of PES. Unlike pooling- and encoder-based approaches—most of which relied on pre-extracted features such as Baidu embeddings \cite{zhou2021feature} on SoccerNet—E2E-Spot was designed as a fully end-to-end trainable architecture built on RegNet-Y \cite{radosavovic2020designing}. To capture fine-grained temporal dynamics with minimal computational overhead, the model incorporates Gate Shift Modules (GSM) \cite{sudhakaran2020gate}, which explicitly model temporal shifts by selectively exchanging feature information across time through a gating mechanism. The sequential features are then processed by a bidirectional GRU \cite{chung2014empirical}, followed by an MLP that outputs per-frame event probabilities. This design enables frame-accurate localization, positioning E2E-Spot as a cornerstone in the development of PES.

Despite the success of E2E-Spot, Tran et al. \cite{tran2024unifying} highlighted its limitation of relying primarily on global temporal modeling. To address this, they introduced the Unifying Global and Local (UGL) module. While retaining the RegNet-Y backbone with GSM for global spatiotemporal encoding, UGL integrates GLIP \cite{li2022grounded}, a vision-language model, to perform fine-grained local semantic reasoning (e.g., recognizing contextual cues such as the presence of a referee or a ball). By combining global temporal context with localized semantic awareness, UGL improves the detection of subtle or ambiguous events, including fouls and off-screen actions, and achieves state-of-the-art performance on the SoccerNet benchmark. However, the reliance on GLIP is both a strength and a limitation: although its pretrained representations transfer well to SoccerNet without requiring fine-tuning, applying UGL to other sports often demands GLIP fine-tuning, which introduces substantial computational overhead and can lead to reduced performance.

One major challenge in PES is frame discriminability, as adjacent frames often appear visually similar with only subtle differences. To improve temporal resolution and disambiguate closely spaced events, Xarles et al. proposed T-DEED \cite{xarles2024t}, a Transformer-based encoder–decoder architecture specifically designed for PES. The model introduces Scalable-Granularity Perception (SGP) layers \cite{shi2023tridet}, originally developed to address the rank-loss problem in Transformers, thereby enhancing token discriminability within sequences. In addition, it incorporates Gate Shift Fusion (GSF) modules \cite{sudhakaran2023gate}, a variant of the GSM module with improved fusion mechanisms prior to shifting, enabling stronger retention of temporal discriminability across tightly clustered frames. This combination proves especially effective for fast-paced sports where events are rapid and visually similar, achieving state-of-the-art performance across multiple PES datasets.

Santra et al. proposed the Adaptive Spatio-Temporal Refinement Module (ASTRM) \cite{santra2025precise}, further advancing PES modeling. ASTRM enhances backbone features by jointly incorporating spatial and temporal cues through three dedicated blocks: local spatial, local temporal, and global temporal. The refined features are then passed into a temporal module consisting of a bidirectional GRU followed by an MLP, which outputs per-frame event classifications.  
To address the severe class imbalance common in PES, the authors also introduced the Soft Instance Contrastive (SoftIC) loss. This loss encourages feature compactness and improves inter-class separability, while resolving a key limitation of the Instance Contrastive Loss (IC Loss) \cite{han2022opendet}. Specifically, IC Loss assumes that each sample has a single label, an assumption that breaks when mixup augmentation \cite{zhang2018mixup} generates samples with mixed labels. SoftIC accounts for the class-specific weights introduced by mixup, enabling more effective learning under imbalanced conditions.

Following this line of research, Xu et al. proposed the Multi-Scale Attention GSM (MSAGSM) \cite{xu2025multi}, which addresses a key limitation of the original GSM—its ability to shift only between adjacent frame features. MSAGSM extends this by enabling feature shifting across longer temporal windows. To improve efficiency, the authors argue that most visual features in sports videos correspond to the background and do not require shifting. They therefore introduce a channel-group attention mechanism that selectively emphasizes informative regions before shifting, enhancing both efficiency and performance. A noted limitation of this approach is its sensitivity to hyperparameters, as the optimal temporal shifting range can vary across different sports.

Overall, frame-aware methods represent the latest and most effective approaches for addressing the PES task. Their key advantage lies in the ability to perform true frame-level classification, enabling precise temporal localization. In contrast to pooling- and encoder-based approaches—which primarily adapt architectures from generic video understanding—frame-aware methods are specifically designed with the characteristics of sports videos in mind. As a result, they currently achieve state-of-the-art performance across multiple sports video event detection benchmarks as shown in Table \ref{tab:as_pes_summary}.

\subsubsection{Multi-Modal Based Methods}
Multi-modal approaches extend purely visual modeling by incorporating complementary modalities, most notably audio. Acoustic cues—such as whistles, ball strikes, or crowd reactions—often align with event boundaries and provide contextual signals that may not be easily inferred from visual frames alone. By fusing modalities, these methods aim to improve robustness and temporal precision in spotting.  

Vanderplaetsen et al. \cite{vanderplaetse2020improved} conducted one of the earliest systematic studies of audio–visual fusion for soccer event detection. They explored early, late, and hybrid fusion strategies for combining audio spectrogram features with visual embeddings. Their results indicated that late fusion—where audio and visual streams are processed independently and only combined before the classification stage—yielded the best performance on SoccerNet. This suggests that modality-specific encoders are more effective at capturing the unique dynamics of each signal, and that overly tight integration (e.g., at the feature extraction stage) may introduce noise.  

Building on this direction, Xarles et al. \cite{xarles2023astra} proposed ASTRA, a Transformer-based encoder–decoder architecture designed to jointly process audio and visual embeddings. Instead of simple concatenation, ASTRA introduces learnable cross-modal queries within a multi-head attention framework, enabling the model to adaptively focus on relevant cues from both streams. For instance, whistles or spikes in crowd noise are aligned with frame-level video representations, allowing the model to highlight ambiguous moments such as fouls or missed shots. This cross-modal reasoning enabled ASTRA to achieve strong performance on SoccerNet, underscoring the potential of attention-based fusion for sports event detection.  

Although multi-modal integration represents a promising research direction, it currently faces several practical challenges. Most available sports datasets contain limited or weakly informative audio, restricting the effectiveness of models that rely on cross-modal signals. Furthermore, in semi-professional, Paralympic, or amateur settings, multiple games are often played simultaneously in shared venues, leading to significant background noise and cross-contamination across matches. In such cases, audio cues may not only provide little benefit but can actively degrade performance if not properly filtered. Consequently, while multi-modal models demonstrate clear advantages under curated broadcast conditions, broader adoption in real-world sports analytics will require improved datasets, robust denoising techniques, and adaptive mechanisms to handle inconsistent or noisy audio streams.

\subsubsection{Other Models}
Despite recent advances, a persistent bottleneck for both AS and PES remains the reliance on large-scale annotated datasets, which are costly and time-consuming to produce. This has motivated research into strategies that reduce annotation requirements or exploit unlabeled data more effectively.  

Giancola et al. \cite{giancola2023towards} proposed the first active learning framework for action spotting to address this challenge. Their pipeline begins with a baseline model trained on a small labeled subset of SoccerNet. The model is then applied to unlabeled videos, producing predictions that are ranked by uncertainty using entropy- and confidence-based heuristics. The most informative samples are selected for manual annotation and iteratively added back into the training pool. This process prioritizes labeling clips that provide the highest information gain, reducing redundant annotation. Experiments demonstrated that their framework achieved competitive results with only one-third of the labeled data required by fully supervised baselines, highlighting the potential of active learning to lower annotation costs in large-scale sports datasets. However, the method still depends on repeated human-in-the-loop annotation cycles, which may limit scalability for rapidly expanding datasets or sports with highly diverse event taxonomies.  

More recently, Denize et al. introduced COMEDIAN \cite{denize2024comedian}, the first AS-specific framework to unify self-supervised learning (SSL) and knowledge distillation (KD) for pretraining spatiotemporal Transformers. The architecture separates modeling into two components: a spatial transformer that captures short-range local context within clips and a temporal transformer that encodes long-range dynamics across sequences. For SSL, the spatial branch is trained with Momentum Contrast (MoCo) \cite{he2020momentum}, encouraging robust representations across temporally adjacent clips. Simultaneously, a Soft Contrastive Loss (SCE) \cite{denize2023similarity} distills knowledge from a feature bank generated by a large pre-trained video model, transferring semantic richness into the AS framework. After this pretraining stage, the model is fine-tuned with labeled data for the AS task, yielding state-of-the-art performance on SoccerNet-v2 while requiring significantly fewer annotations. This demonstrates the promise of combining SSL and KD to bootstrap event spotting models from unlabeled video corpora.

These methods highlight an emerging shift towards data-efficient learning in AS and PES. Active learning frameworks reduce annotation redundancy by selectively labeling the most informative clips, while SSL + KD approaches leverage vast amounts of unlabeled video to pretrain strong representations. However, challenges remain: active learning pipelines are still annotation-intensive and require careful design of selection heuristics, while SSL and KD approaches are heavily dependent on the choice of pre-trained models and may inherit their biases. Furthermore, the diversity of sports poses an additional barrier, as strategies effective in soccer may not directly transfer to domains with scarcer data or different event semantics. Nevertheless, reducing annotation reliance remains a crucial step toward scaling PES systems to broader sports contexts, particularly outside well-curated professional broadcast datasets.

\section{Datasets}\label{sec:datasets}

Datasets play a critical role in supervised deep learning, providing the foundation for both model training and evaluation. Transformer-based architectures \cite{arnab2021vivit, liu2021swin, dosovitskiy2020image} are particularly data-dependent, often requiring large-scale, high-quality datasets to achieve strong generalization. However, annotating sports videos remains a time-intensive and expertise-driven process. For example, distinguishing between different serve types in table tennis or tennis can be highly challenging due to subtle motion variations and the high speed of play \cite{wu2022survey}. Consequently, high-quality, precisely annotated datasets are especially valuable, as sports actions often exhibit limited generalization across different contexts.

In this section, we review publicly available sports-related datasets commonly used for event detection, grouping them by sport genre. For each dataset, we provide a detailed description and discuss its current limitations. A summary of these datasets is presented in Table~\ref{tab:sports_datasets_summary}.

\begin{table}[ht]
\centering
\renewcommand{\arraystretch}{1.3}
\caption{Overview of sports-related datasets for event detection. 
\textit{Spotting} denotes precise frame-level annotations, while \textit{Interval} annotations specify action start and end times.}
\LARGE
\resizebox{\linewidth}{!}{%
\begin{tabular}{l|c|c|c|c|c}
\toprule
\textbf{Dataset} & \textbf{Sport} & \textbf{Year} & \textbf{Size / Duration} & \textbf{Annotation Type} & \textbf{Categories / Events} \\
\midrule
SoccerNet \cite{giancola2018soccernet} & Soccer & 2018 & 500 videos / 764 hrs & Spotting & 3 (goals, cards, substitutions) \\
SSET \cite{feng2020sset} & Soccer & 2020 & 350 videos / 282 hrs & Interval & 11 event types, 15 story types \\
SoccerDB \cite{jiang2020soccerdb} & Soccer & 2020 & 346 videos / 669 hrs & Interval & 10 \\
SoccerNet-v2 \cite{deliege2021soccernet} & Soccer & 2021 & 500 videos / 764 hrs & Spotting & 17 event classes \\
SoccerNet Ball AS \cite{cioppa2024soccernet} & Soccer & 2023 & 7 videos & Spotting & 12 ball-action events \\
Tenniset \cite{faulkner2017tenniset} & Tennis & 2017 & 5 videos & Interval & 6 \\
Tennis \cite{hong2022spotting} & Tennis & 2022 & 3,345 clips & Spotting & 6 (court-specific ball contacts) \\
OpenTTGames \cite{voeikov2020ttnet} & Table Tennis & 2020 & 12 videos & Spotting & 3 \\
P$^2$A \cite{bian2024p2anet} & Table Tennis & 2024 & 2,721 videos / 272 hrs & Interval & 14 fine-grained / 8 high-level stroke classes \\
TTA \cite{xu2025multi} & Table Tennis & 2025 & 39 videos & Spotting & 8 \\
NCAA \cite{ramanathan2016detecting} & Basketball & 2016 & 257 videos / ~1.5 hrs each & Interval & 14 (e.g., 3-point, dunk, steal) \\
Badminton Olympic \cite{ghosh2018towards} & Badminton & 2018 & 27 videos & Interval & 12 \\
Figure Skating \cite{hong2021video} & Figure Skating & 2021 & 11 videos & Interval & 4 transitions \\
FineDiving \cite{xu2022finediving} & Diving & 2022 & 300 videos & Interval & 52 key pose transitions \\
FineGym \cite{shao2020finegym} & Gymnastics & 2020 & 5,374 videos & Spotting & 32 fine-grained actions \\
MCFS \cite{liu2021temporal} & Figure Skating & 2021 & 11,656 segments / 17.3 hrs & Interval & 130 across 4 event sets \\
\bottomrule
\end{tabular}}
\label{tab:sports_datasets_summary}
\end{table}

\subsection{Soccer}

\textbf{SoccerNet-V1} \cite{giancola2018soccernet} was the first large-scale benchmark for sports video analysis, covering multiple tasks including event detection. It contains 500 full-match broadcasts (764 hours, ~4TB) from major European championships (2015–2017). Events are annotated from official match reports with one-second resolution for three event types. While it supports both AS and PES tasks—for example, the “card” label marks the moment a referee issues a booking—the coarse and ambiguous one-second annotations limit temporal precision. As a result, most early methods developed on SoccerNet-V1 focused on AS rather than PES.

\textbf{SSET} \cite{feng2020sset} is a smaller dataset relative to SoccerNet, containing 350 videos covering multiple soccer games, totaling 282 hours. It defines 11 event types and 15 story types. Designed primarily for TAL, its event annotations are interval-based. For instance, a "kick" event is annotated from the moment a key player prepares to kick until the ball lands or exits the field. 

\textbf{SoccerDB} \cite{jiang2020soccerdb} contains 346 high-quality soccer match videos, incorporating 270 matches from SoccerNet and 76 matches from the Chinese Super League (2017–2018) and FIFA World Cup editions. The dataset occupies 1.4TB and has a total duration of 668.6 hours. It defines 10 soccer event types with clear temporal boundaries, making it highly suitable for event detection tasks.

\textbf{SoccerNet-v2} \cite{deliege2021soccernet} extends SoccerNet by expanding the number of action classes from 3 to 17, introducing more detailed events such as “Foul,” “Throw-in,” and “Shot on target.” The most significant change compared to SoccerNet-V1 is that each event is annotated with a single timestamp rather than a one-second interval. In addition, each timestamp is assigned a visibility tag indicating whether the action is explicitly visible or inferred, which introduces further challenges for automated detection. This design enables the PES task and also allows evaluation of whether models leverage broader temporal context to understand the game, or merely rely on local spatial cues.

\textbf{SoccerNet Ball Action Spotting} \cite{cioppa2024soccernet} extended SoccerNet-v2 to focus on fine-grained ball interactions, requiring frame-level precision for frequent events. It initially annotated “pass” and “drive” actions across 7 matches (11,041 labels), and was later expanded in 2024 to include 12 ball-related classes, supporting detailed modeling of gameplay flow.

\subsection{Racket Sports}

\textbf{Tenniset} \cite{faulkner2017tenniset} consists of five full-match videos from the 2012 London Olympic Games, sourced from YouTube. It defines six event types, such as "set," "hit," and "serve," annotated with precise temporal intervals. In addition, Tenniset provides textual descriptions of actions, such as "quick serve is an ace," enabling multimodal learning that combines video and text modalities.

The \textbf{Tennis} dataset \cite{hong2022spotting}, built upon the Vid2Player dataset \cite{zhang2021vid2player}, comprises 3,345 video clips from 28 professional tennis matches recorded at 25 or 30 FPS. Events are categorized into six classes, including "player serve ball contact," "regular swing ball contact," and "ball bounce," further divided based on court side (near or far court). The dataset supports fine-grained action spotting in tennis and facilitates evaluation under strict temporal precision settings.

\textbf{OpenTTGames} \cite{voeikov2020ttnet} consists of 12 high-definition table tennis matches recorded at 120 FPS, containing 4,271 labeled events. The dataset defines three event types—ball bounces, net hits, and empty events—all annotated with frame-level precision. OpenTTGames is particularly suited for training models on bounce detection under high-speed gameplay conditions.  

\textbf{P$^2$A} \cite{bian2024p2anet} is a large-scale table tennis dataset comprising 2,721 broadcast videos (272 hours) from major tournaments. It includes 14 fine-grained stroke classes grouped into 8 higher-level action categories, with frame-level annotations validated by professionals, making it one of the most comprehensive stroke-level benchmarks.  

\textbf{TTA} \cite{xu2025multi} represents the latest table tennis PES benchmark, consisting of 39 para-professional matches. Unlike broadcast-only datasets, TTA captures real-world recording conditions with non-ideal camera angles, frequent occlusions, and less controlled environments. It is the first benchmark to target PES in para-sport contexts, making it highly relevant for practical and inclusive sports analytics.  

\textbf{Badminton Olympic} \cite{ghosh2018towards} contains 27 badminton match videos sourced from the official Olympic YouTube channel. It includes time-interval annotations for 12 action types, such as "serve," "backhand," and "smash," as well as point-level annotations, making it suitable for both action spotting and match-level analysis.

\textbf{BadmintonTrack} \cite{sun2020tracknetv2} is another badminton dataset, comprising 77,000 annotated frames from 26 unique singles matches filmed from an overhead broadcast-view camera. Originally, timestamp information indicating when a player struck the shuttlecock was included \cite{liu2022monotrack}, although this metadata has since been removed in updated versions.

\subsection{Other Sports}

\textbf{NCAA} \cite{ramanathan2016detecting} consists of 257 untrimmed college basketball game videos, each approximately 1.5 hours long. The dataset provides 14,548 video segments, with precise start and end times for 14 action categories, supporting temporal action localization tasks.

The \textbf{Figure Skating} dataset \cite{hong2021video} contains 11 videos recorded at 25 FPS, featuring 371 short program performances from the 2010–2018 Winter Olympics and the 2017–2019 World Championships. It defines four transition event types critical for evaluating temporal precision in figure skating analysis.

\textbf{FineDiving} \cite{xu2022finediving} comprises 300 professional diving videos collected from major international competitions, including the Olympics, World Cup, and World Championships. It defines 52 fine-grained action types, 29 sub-action types, and 23 difficulty levels, making it a comprehensive benchmark for procedural action quality assessment. Although the original annotations were designed for action quality evaluation rather than temporal spotting, Hong et al. \cite{hong2022spotting} later adapted the dataset to support the PES task by refining frame-level event annotations.

\textbf{FineGym} \cite{shao2020finegym} provides 5,374 gymnastics performances from international competitions. Each video is annotated with a hierarchical structure categorizing 32 spotting classes, covering disciplines such as balance beam and floor exercise, enabling fine-grained action spotting and classification.

\textbf{MCFS} \cite{liu2021temporal} is a large-scale figure skating dataset comprising 11,656 video segments across 38 competitions, totaling 17.3 hours and 1.7 million frames. Annotations follow a hierarchical structure of 4 high-level action sets, 22 subsets, and 130 element actions, making MCFS highly suitable for dense temporal action localization tasks in figure skating.

\subsection{Limitations}  
A common limitation across most sports datasets is their reliance on professional broadcast footage. While such data provide high video quality and consistent coverage, they do not reflect everyday contexts such as semi-professional, youth, para-sport, or amateur matches, where camera placement, video quality, and gameplay dynamics differ substantially. Consequently, models trained on these datasets may struggle to generalize or transfer effectively to less controlled, real-world scenarios.

\section{Evaluation Metrics}\label{sec:evaluation_metrics}

Sports video event detection employs different evaluation metrics depending on the task—TAL, AS, or PES—each measuring distinct aspects of temporal localization and classification. For detailed mathematical derivations, readers are referred to the Supplementary Materials.

\subsection{Mean Average Precision (mAP@T-IoU)}
For TAL, the standard evaluation metric is mean Average Precision computed with temporal Intersection over Union thresholds (mAP@T-IoU). Predictions are considered true positives if their Temporal IoU (T-IoU) with the ground truth exceeds a given threshold.

\begin{equation}
\text{Precision} = \frac{TP}{TP + FP}, \quad
\text{Recall} = \frac{TP}{\text{Total GT}},
\end{equation}

\begin{equation}
\text{T-IoU} = \frac{|I_p \cap I_g|}{|I_p \cup I_g|},
\end{equation}
where $I_p$ and $I_g$ denote the predicted and ground-truth temporal intervals, respectively.

Average Precision (AP) is computed for each class, and mAP is calculated by averaging across all classes:

\begin{equation}
\mathrm{mAP} = \frac{1}{C} \sum_{c=1}^{C} \mathrm{AP}_c.
\label{math:map}
\end{equation}
where \( C \) is the total number of action classes and \( \mathrm{AP}_c \) is the Average Precision computed for the \( c^{\text{th}} \) class.
Though standard, mAP is highly sensitive to T-IoU thresholds and may overly penalize minor misalignments in sports scenarios with ambiguous boundaries.

\subsection{AR@AN and AUC}
For TAPG, AR@AN evaluates how many ground-truth segments are recovered given a fixed number of proposals per video. The Area Under the Curve (AUC) measures average recall across varying proposal counts:

\begin{equation}
\mathrm{AUC} = \int_{0}^{N} \text{AR}(n) \, dn.
\end{equation}
where \( \text{AR}(n) \) is the average recall when using \( n \) proposals, and \( N \) is the maximum number of proposals considered. 
These metrics emphasize proposal coverage but ignore redundancy and precision.

\subsection{Tolerance Windows and mAP@\texorpdfstring{$\delta$}{delta}}
For AS and PES, mAP is the primary evaluation metric. It is computed under a temporal tolerance window $\delta$ around the ground-truth timestamp (e.g., $\delta = 5$–$60$ frames for AS, $\delta = 0$–$2$ frames for PES). This is denoted as mAP@$\delta$ to distinguish it from mAP@T-IoU used in TAL.

AP is computed per class by first ranking predictions according to confidence scores and then integrating the resulting Precision–Recall (PR) curve. Formally, for class $c$ with $N_c$ ranked predictions, AP is given by:  
\begin{equation}
\mathrm{AP}_c^\delta = \sum_{i=1}^{N_c} \big(\mathrm{Rec}_c(i)-\mathrm{Rec}_c(i-1)\big)\,\mathrm{Prec}_c(i),
\end{equation}
where $\mathrm{Prec}_c(i)$ and $\mathrm{Rec}_c(i)$ denote the precision and recall after considering the top-$i$ predictions.  
The overall mean Average Precision is then obtained by averaging over all classes:
\begin{equation}
\mathrm{mAP}@\delta = \frac{1}{C}\sum_{c=1}^{C} \mathrm{AP}_c^\delta.
\end{equation}

\textbf{Limitation.} A key limitation of mAP@$\delta$ in the PES setting is that contradictory predictions at the same frame are not consistently penalized. In practice, prediction thresholds are often set very low (e.g., 0.1), which allows multiple classes to be retained for a single frame. Since AP is computed independently per class, any extra prediction for a class with \emph{no ground-truth events in that sequence} is simply ignored rather than counted as a false positive. Moreover, evaluation toolkits handle this situation inconsistently: some exclude classes with no ground-truth from the mAP average (so spurious predictions have no effect), while others assign them an AP of zero (which penalizes the model). This inconsistency makes reported mAP scores difficult to interpret and compare across implementations.

For example, consider a table tennis frame $x$ annotated only as \textit{stroke} ($y_{stroke}(x)=1$, $y_{serve}(x)=0$). Suppose the model outputs:
\[
\hat{p}_{stroke}(x) = 0.3, \qquad \hat{p}_{serve}(x) = 0.4.
\]
In AP computation:  
- For the \textit{stroke} class, $\hat{p}_{stroke}(x)$ is matched to the ground truth and counted as a true positive.  
- For the \textit{serve} class, since there are no ground-truth serve events in this sequence, $\hat{p}_{serve}(x)$ is ignored; it does not enter into the precision–recall calculation and is not treated as a false positive.  

As a result,  
\[
\mathrm{AP}_{stroke} = 1, \quad \mathrm{AP}_{serve} \text{ is excluded or left unaffected}.
\]
The overall mAP remains artificially high despite the contradictory prediction (\textit{stroke} + \textit{serve}) at the same frame. This behavior stems from the metric’s multi-label origins and favors over-predictive systems, even though in sports domains only one event can occur per timestamp.

\textbf{Proposed Modification.}  
We recommend stricter benchmarking protocols that enforce \emph{top-1 filtering}, where only the highest-scoring class is retained per frame, and compute AP by sweeping over confidence thresholds rather than ranking predictions. This approach penalizes extra predictions and provides an evaluation that more faithfully reflects real deployment requirements.

\paragraph{Top-1 per frame.}  
For each frame $f$, with class scores $s_{f,c}$:
\[
\hat{c}_f=\arg\max_c s_{f,c}, \qquad \hat{s}_f=\max_c s_{f,c}.
\]

\paragraph{AP via threshold sweep.}  
With top-1 filtering applied, each frame contributes at most one prediction. Varying the confidence threshold $\tau$ from high to low traces the PR curve in the standard way. The AP for class $c$ is then:
\[
\mathrm{AP}_c^\delta=\sum_{k=1}^{K} \big(\mathrm{Rec}_c(\tau_k)-\mathrm{Rec}_c(\tau_{k+1})\big)\,\mathrm{Prec}_c(\tau_k),
\]
where $\tau_1>\tau_2>\cdots>\tau_K$ are the distinct confidence thresholds (or a fixed grid).  

\paragraph{Final metric.}  
\[
\mathrm{mAP}@\delta=\frac{1}{C}\sum_{c=1}^{C}\mathrm{AP}_c^{\delta}.
\]

This stricter protocol (i) enforces one class per frame, (ii) penalizes over-prediction, and (iii) evaluates recall effectiveness by integrating precision over confidence thresholds instead of intra-frame ranking.

\section{Practical Applications}\label{sec:app}
\begin{figure}[htbp]
    \centering
    \includegraphics[width=1\linewidth]{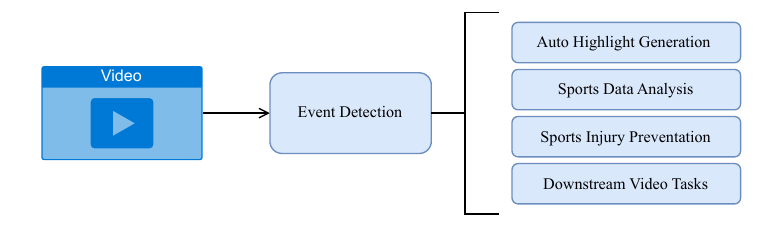}
    \caption{Overview of practical applications enabled by sports video event detection.}
    \Description{A conceptual diagram illustrating four main applications of sports video event detection: automatic highlight generation, sports data analysis, injury prevention, and downstream video tasks. Each application is shown as a labeled block connected to the central event detection component.}
    \label{fig:app}
\end{figure}

Sports video event detection enables practical benefits across media, performance analysis, and athlete health. By structuring raw footage into meaningful events, these systems support highlight generation, tactical evaluation, efficient video processing, and injury prevention. The following subsections outline key applications. A summary of areas covered in this section is shown in Figure \ref{fig:app}.

\subsection{Automatic Game Highlight Generation}
By automatically detecting key moments such as goals, fouls, or ball bounces, event detection facilitates efficient content indexing and retrieval \cite{cioppa2024soccernet}. Broadcasters and media platforms can then generate highlight reels in real time, reducing manual effort while ensuring that audiences capture all significant events. This capability is especially valuable for large-scale tournaments, where vast amounts of footage must be processed quickly and accurately.

\subsection{Sports Video Data Analysis}
Another major application is performance analysis for athletes and coaches. Automated detection of fine-grained events, such as specific strokes in racket sports or tactical plays in team sports, enables precise breakdowns of strategies and player behaviors. Recording points won and linking them to the events that caused them is also critical for tactical analysis. These insights allow coaches to deliver targeted feedback, while athletes benefit from real-time, data-driven evaluations that can be provided during breaks in a match.

\subsection{Efficient Video Handling for Downstream Tasks}
Beyond direct applications, event detection also serves as an efficient preprocessing step for other computer vision tasks. Instead of processing entire matches frame by frame, detected events can act as temporal anchors that highlight only the most informative segments. For example, rather than tracking the ball continuously throughout a match, tracking algorithms can be applied only around bounce events where precision matters most. Similarly, action recognition systems can be guided by event detectors to focus on short clips surrounding serves, enabling more accurate classification of serve types without excessive computation. This targeted handling of video data not only reduces processing costs but also improves the effectiveness of downstream tasks such as player behavior analysis, tactical modeling, and strategy discovery.

\subsection{Injury Prevention and Workload Monitoring}
Event detection can also play an important role in safeguarding athlete health. By recognizing repetitive micro-events such as jumps, sprints, or strokes, systems can automatically quantify training and match workloads. This information provides sports scientists and medical staff with objective measures of player exertion, helping to prevent overuse injuries. For example, detecting abnormal movement patterns or sudden increases in workload can serve as early warning signals for potential injuries. Furthermore, long-term monitoring of event-level data enables personalized training programs, ensuring that athletes maintain peak performance while minimizing health risks. Such applications are particularly valuable in elite sports, where even small improvements in injury prevention can have significant impacts on team success and athlete longevity.

\section{Challenges and Future Directions} \label{sec:challenges_directions}
In this section, we critically examine current challenges in sports event detection and outline specific, actionable future research directions to address these limitations.

\subsection{Generalization Across Diverse Sports}
While many AS and PES models achieve strong results within individual sports—particularly soccer, given the scale of SoccerNet-V1 and SoccerNet-V2—they often rely heavily on domain-specific visual and contextual cues such as camera angles, common action semantics, and gameplay structure. This reliance limits transferability to sports with different visual dynamics, motion patterns, and temporal scales.

A core limitation in current approaches is the dependence on backbone architectures originally developed for image classification or coarse-grained action recognition. These architectures typically process video in fixed-length chunks and aggregate features spatially and temporally, which suppresses subtle frame-level distinctions crucial for spotting tasks. In contrast, PES requires temporally fine-grained representations that can capture minimal variations between adjacent frames—such as a foot making contact with a ball or a player crossing a boundary line.

To improve generalization and robustness, future work should prioritize frame-level representation learning tailored to the demands of spotting tasks. Promising directions include:
\begin{itemize}
\item Developing encoders that preserve local temporal granularity, using lightweight 1D CNNs, temporal contrastive learning, or frame-attentive modules to enhance discriminative capacity.
\item Leveraging multimodal pretraining (e.g., CLIP \cite{radford2021learning}) to align visual, textual, and audio cues into semantically rich frame embeddings suitable for cross-sport transfer.
\item Exploring adaptive frame sampling strategies that focus representational capacity on moments of high temporal importance, improving both efficiency and localization accuracy.
\end{itemize}

By enhancing frame-wise representation learning, future AS and PES models will be better equipped to generalize across diverse sports scenarios, achieving higher temporal precision while reducing reliance on domain-specific heuristics.

\subsection{Unsupervised and Low-Supervision Methods}
Creating large-scale labeled datasets for sports event detection is costly, labor-intensive, and often requires expert knowledge, particularly in technical sports such as gymnastics, tennis, and figure skating. To mitigate these barriers, recent work has explored low-supervision paradigms such as knowledge distillation and active learning \cite{denize2024comedian, giancola2023towards}, which reduce reliance on extensive annotations by transferring knowledge from pretrained models or selectively labeling the most informative samples.

Fully unsupervised and self-supervised approaches, however, remain in their infancy. Future research directions include:
\begin{itemize}
\item Designing self-supervised frameworks that exploit temporal consistency, contrastive objectives, or multimodal alignment to learn meaningful event representations from unlabeled or weakly labeled sports videos.
\item Combining unsupervised learning with domain adaptation to improve generalization across sports with diverse visual dynamics and gameplay structures.
\end{itemize}

Advancing in these directions will be critical for building scalable, efficient, and widely applicable event detection models, especially in sports where annotations are scarce or costly.

\subsection{Enhanced Multimodal Fusion Approaches}
Although most existing AS and PES methodologies rely primarily on visual data, audio cues can substantially enrich the detection of critical events in sports, as demonstrated by \cite{xarles2023astra}. Examples include ball impact sounds, crowd reactions, or figure skaters landing on ice—all of which provide complementary temporal signals.  

Current multimodal models largely adopt simple fusion strategies, such as concatenation or late fusion \cite{vanderplaetse2020improved, xarles2023astra}, which fail to capture the complex interactions between modalities. Moreover, in many general-level sports, audio data is often unavailable or dominated by noise (e.g., background music or commentary), unlike in elite-level broadcasts where clean signals are more common. To overcome these limitations, future research should explore more advanced fusion techniques and robust noise-handling strategies.  

\begin{itemize}
\item Attention-based cross-modal transformers and gated attention mechanisms that dynamically weight and integrate audio–visual cues.  
\item Modality-specific encoders combined with temporal alignment mechanisms to capture precise event timings in noisy or visually ambiguous settings.  
\item Leveraging commentary audio through automatic speech recognition and natural language processing to provide additional semantic context and weak supervision signals, aligning spoken descriptions with visual events.  
\item Noise-robust feature extraction and denoising strategies to improve the reliability of audio cues in non-professional or crowd-sourced sports footage.  
\end{itemize}

Advancing multimodal fusion methodologies represents a key opportunity to enhance the accuracy, robustness, and practical applicability of AS and PES systems in sports video analytics.

\subsection{Real-World Applications: Gaps in Datasets and Evaluation Protocols}

Despite impressive progress in AS and PES research, a substantial gap remains between academic benchmarks and real-world deployment. Most existing datasets are curated from professional broadcasts, captured with high-quality cameras, ideal lighting, and fixed angles. While this consistency supports reliable annotations and evaluation, it fails to reflect the realities faced by analysts, coaches, and practitioners outside elite or televised contexts. At amateur or semi-professional levels, footage is often self-recorded using handheld devices or static single-angle setups under suboptimal conditions, where models trained on curated datasets may struggle to generalize.

Evaluation protocols present similar limitations. Current benchmarks often compute mean mAP with low confidence thresholds (e.g., 0.1), which allows multiple class predictions per frame. In PES, however, a single frame almost never contains more than one event. While multi-label predictions boost recall and improve mAP scores, they provide an inflated view of performance and are misaligned with practical needs. This issue is especially evident in racket sports, where only one event (e.g., hit or bounce) can occur at a given time, and over-prediction directly reduces usefulness for tasks like rule enforcement or tactical feedback.

To bridge these gaps, future work should:
\begin{itemize}
    \item Create and evaluate datasets recorded in unconstrained, real-world environments to ensure robustness beyond broadcast-quality footage.
    \item Establish evaluation protocols that penalize over-prediction and reward frame-level discriminability, such as top-1 class selection or calibrated confidence thresholds aligned with deployment requirements.
    \item Release datasets spanning diverse venues, competition levels, and camera setups to reduce domain gaps between research and practice.
\end{itemize}

Closing these gaps is essential for building AS and PES systems that are not only accurate on benchmarks but also reliable, efficient, and trustworthy in practice.

\section{Conclusion}\label{sec:conclusion}
In this survey, we reviewed deep learning-based methods, datasets, and evaluation protocols for video event detection, with a particular focus on TAL, AS, and PES in sports analytics. We highlighted several key challenges, including evaluation protocols that do not fully account for multiple predictions, the underrepresentation of datasets covering the broader sports community, the limited generalizability of methods across different sports, the heavy reliance on extensive annotations, and the underutilization of multimodal cues.

To address these gaps, future research should emphasize frame-level models with task-specific backbones, robust cross-sport evaluation, and scalable learning paradigms such as self-supervised and active learning. Incorporating multimodal signals—including visual, audio, and text—also has strong potential to enhance temporal precision and contextual understanding.

Addressing these challenges will pave the way for more accurate, generalizable, and efficient sports video event detection systems.


\bibliographystyle{ACM-Reference-Format}
\bibliography{references}

\end{document}